\documentclass[runningheads]{llncs}
\usepackage{caption}
\usepackage{subcaption}
\usepackage{amsfonts}
\usepackage{amsmath}
\usepackage{amssymb}
\usepackage{bm}
\usepackage{graphicx}
\usepackage{hyperref}
\usepackage{biblatex}
\usepackage{pifont}
\usepackage{booktabs}
\usepackage{color}
\begin{document}
\title{Regular SE(3) Group Convolutions for Volumetric Medical Image Analysis}
%
%
\author{Thijs P. Kuipers\orcidID{0009-0007-7198-2856} \and Erik J. Bekkers\orcidID{0000-0003-4418-2160}}

%
\authorrunning{Thijs P. Kuipers \and Erik J. Bekkers}
%
\institute{Amsterdam Machine Learning Lab, Informatics Institute, University of Amsterdam}
\maketitle              
\begin{abstract}
Regular group convolutional neural networks (G-CNNs) have been shown to increase model performance and improve equivariance to different geometrical symmetries. This work addresses the problem of SE(3), i.e., roto-translation equivariance, on volumetric data. Volumetric image data is prevalent in many medical settings. Motivated by the recent work on separable group convolutions, we devise a SE(3) group convolution kernel separated into a continuous SO(3) (rotation) kernel and a spatial kernel. We approximate equivariance to the continuous setting by sampling uniform SO(3) grids. Our continuous SO(3) kernel is parameterized via RBF interpolation on similarly uniform grids. We demonstrate the advantages of our approach in volumetric medical image analysis. Our SE(3) equivariant models consistently outperform CNNs and regular discrete G-CNNs on challenging medical classification tasks and show significantly improved generalization capabilities. Our approach achieves up to a 16.5\% gain in accuracy over regular CNNs.
\end{abstract}

\section{Introduction}

Invariance to geometrical transformations has been long sought-after in the field of machine learning \cite{cohen2013, kondor}. The strength of equipping models with inductive biases to these transformations was shown by the introduction of convolutional neural networks (CNNs) \cite{lecun1995convolutional}. Following the success of CNNs, \cite{origingcn} generalized the convolution operator to commute with geometric transformation groups other than translations, introducing group-convolutional neural networks (G-CNNs), which have been shown to outperform conventional CNNs \cite{3dsteerable, 3dgcn1, bekkers2020bspline, knigge}.

Early G-CNNs were mainly concerned with operating on 2D inputs. With the increase in computing power, G-CNNs were extended to 3D G-CNNs. Volumetric data is prevalent in many medical settings, such as in analyzing protein structures \cite{deeprank1} and medical image analysis \cite{medical1, medical2, medmnistv1, medmnistv2}. Equivariance to symmetries such as scaling and rotations is essential as these symmetries often naturally occur in volumetric data. Equivariance to the group of 3D rotations, SO(3), remains a non-trivial challenge for current approaches due to its complex structure and non-commutative properties \cite{3dsteerable}.

An important consideration regarding 3D convolutions that operate on volumetric data is overfitting. Due to the dense geometric structure in volumetric data and the high parameter count in 3D convolution kernels, 3D convolutions are highly susceptible to overfitting \cite{qi}. G-CNNs have been shown to improve generalization compared to CNNs \cite{worrall2017harmonic, 3dgcn1}. However, G-CNNs operating on discrete subgroups can exhibit overfitting to these discrete subgroups \cite{bekkers2018roto}, failing to obtain equivariance on the full continuous group. This effect is amplified for 3D G-CNNs, limiting their improved generalization capabilities.

\subsubsection{Contributions.}

In this work, we introduce regular continuous group convolutions equivariant to SE(3), the group of roto-translations. Motivated by the work on separable group convolutions \cite{knigge}, we separate our SE(3) kernel in a continuous SO(3) and a spatial convolution kernel. We randomly sample discrete equidistant SO(3) grids to approximate the continuous group integral. The continuous SO(3) kernels are parameterized via radial basis function (RBF) interpolation on a similarly equidistantly spaced grid. We evaluate our method on several challenging volumetric medical image classification tasks from the MedMNIST \cite{medmnistv1, medmnistv2} dataset. Our approach consistently outperforms regular CNNs and discrete SE(3) subgroup equivariant G-CNNs and shows significantly improved generalization capabilities. To this end, this work offers the following contributions.

\begin{enumerate}
    \item We introduce separable regular SE(3) equivariant group convolutions that generalize to the continuous setting using RBF interpolation and randomly sampling equidistant SO(3) grids.
    \item We show the advantages of our approach on volumetric medical image classification tasks over regular CNNs and discrete subgroup equivariant G-CNNs, achieving up to a 16.5\% gain in accuracy over regular CNNs.
    \item Our approach generalizes to SE($n$) and requires no additional hyperparameters beyond setting the kernel and sample resolutions.
    \item We publish our SE(3) equivariant group convolutions and codebase for designing custom regular group convolutions as a Python package.\footnote{Our codebase can be accessed at: \url{https://github.com/ThijsKuipers1995/gconv}.}
    
\end{enumerate}

\subsubsection{Paper Outline.}

The remainder of this paper is structured as follows. Section \ref{sec:literature} provides an overview of current research in group convolutions. Section \ref{sec:method} introduces the group convolution theory and presents our approach to SE(3) equivariant group convolutions. Section \ref{sec:results} presents our experiments and an evaluation of our results. We give our concluding remarks in Section \ref{sec:conclusion}.
\section{Literature Overview} \label{sec:literature}

Since the introduction of the group convolutional neural network (G-CNN), research in G-CNNs has grown in popularity due to their improved performance and equivariant properties over regular CNNs. Work on G-CNNs operating on volumetric image data has primarily been focused on the 3D roto-translation group SE(3) \cite{worrall2018cubenet, 3dgcn1, 3dsteerable}. CubeNet was the first introduced 3D G-CNN, operating on the rotational symmetries of the 3D cube \cite{worrall2018cubenet}. The approach presented in \cite{3dgcn1} similarly works with discrete subsets of SE(3). These approaches are not fully equivariant to SE(3). Steerable 3D G-CNNs construct kernels through a linear combination of spherical harmonic functions, obtaining full SE(3) equivariance \cite{3dsteerable}. Other approaches that are fully SE(3) equivariant are the Tensor-Field-Network \cite{tensornet} and N-Body networks \cite{sifre2013rotation}. However, these operate on point clouds instead of 3D volumes.
\section{Separable SE($n$) Equivariant Group Convolutions} \label{sec:method}

This work introduces separable SE(3) equivariant group convolutions. However, our framework generalizes to SE($n$). Hence, we will describe it as such. Section \ref{sec:groupconv} presents a brief overview of the regular SE($n$) group convolution. Section \ref{sec:se3conv} introduces our approach for applying this formulation to the continuous domain.

\subsection{Regular Group Convolutions} \label{sec:groupconv}

The traditional convolution operates on spatial signals, i.e., signals defined on $\mathbb{R}^n$. Intuitively, one signal (the kernel) is slid across the other signal. That is, a translation is applied to the kernel. From a group-theoretic perspective, this can be viewed as performing the group action from the translation group. The convolution operator can then be formulated in terms of the group action. By commuting to a group action, the group convolution produces an output signal that is equivariant to the transformation imposed by the corresponding group.

\subsubsection{The SE($n$) Group Convolution Operator.}

Instead of operating on signals defined on $\mathbb{R}^n$, SE($n$)-convolutions operate on signals defined on the group SE$(n)=\mathbb{R}^n \rtimes \text{SO}(n)$. Given an $n$-dimensional rotation matrix $\mathbf{R}$, and SE(3)-signals $f$ and $k$, the SE($n$) group convolution is defined as follows:
\begin{align}\label{eq:gconv}
    (f *_{\text{group}} k)(\mathbf{x},\mathbf{R}) 
    &=\int_{\mathbb{R}^n}\int_{SO(n)}k\left(\mathbf{R}^{-1}(\Tilde{\mathbf{x}} - \mathbf{x}),\mathbf{R}^{-1}\Tilde{\mathbf{R}}\right)f(\Tilde{\mathbf{x}},\Tilde{\mathbf{R}}){\rm d}\Tilde{\mathbf{R}} {\rm d}\Tilde{\mathbf{x}}.
\end{align}

\subsubsection{The Lifting Convolution Operator.}

Input data is usually not defined on SE$(n)$. Volumetric images are defined on $\mathbb{R}^3$. Hence, the input signal should be lifted to SE$(n)$. This is achieved via a lifting convolution, which accepts a signal $f$ defined on $\mathbb{R}^n$ and applies a kernel $k$ defined on SO$(n)$, resulting in an output signal on SE$(n)$. The lifting convolution is defined as follows:
\begin{align}\label{eq:liftingconv}
    (f *_{\text{lifting}} k)(\mathbf{x})  
    &= \int_{\mathbb{R}^n}k(\mathbf{R}^{-1}(\Tilde{\mathbf{x}} - \mathbf{x}))f(\mathbf{\Tilde{x}}){\rm d}\mathbf{\Tilde{x}}.
\end{align}
The lifting convolution can be seen as a specific case of group convolution where the input is implicitly defined on the identity group element.

\subsection{Separable SE($n$) Group Convolution} \label{sec:se3conv}

The group convolution in Equation \ref{eq:gconv} can be separated into a convolution over $\mathbf{R}$ followed by a convolution over $\mathbb{R}^n$ by assuming $k_{\text{SE}(n)}(\mathbf{x}, \mathbf{R}) = k_{\text{SO}(n)}(\mathbf{R})k_{\mathbb{R}^n}(\mathbf{x})$. This improves performance and significantly reduces computation time \cite{knigge}. 

\subsubsection{The Separable SE($n$) Kernel.}

Let $i$ and $o$ denote in the input and output channel indices, respectively. We separate the SE(3) kernel as follows:
\begin{align}
    k_{\text{SE}(n)}^{io}(\mathbf{x}, \mathbf{R}) = k_{\text{SO}(n)}^{io}(\mathbf{R})k_{\mathbb{R}^n}^{o}(\mathbf{x}).
\end{align}
Here, $k_{\text{SO}(n)}$ performs the channel mixing, after which a depth-wise separable spatial convolution is performed. This choice of separation is not unique. The channel mixing could be separated from the SO($n$) kernel. However, this has been shown to hurt model performance \cite{knigge}.

\subsubsection{Discretizing the Continuous SO($n$) Integral.}

The continuous group integral over SO($n$) in Equation \ref{eq:gconv} can be discretized by summing over a discrete SO($n$) grid. By randomly sampling the grid elements, the continuous group integral can be approximated \cite{wu2019pointconv}. However, randomly sampled kernels may not capture the entirety of the group manifold. This will result in a noisy estimate. Therefore, we constrain our grids to be uniform, i.e., grid elements are spaced equidistantly. Similarly to the authors of \cite{bekkers2020bspline}, we use a repulsion model to generate SO($n$) grids of arbitrary resolution.

\subsubsection{Continuous SO$(n)$ Kernel with Radial Basis Function Interpolation.}

The continuous SO($n$) kernel is parameterized via a similarly discrete SO($n$) uniform grid. Each grid element $\mathbf{R}_i$ has corresponding learnable parameters $\mathbf{k}_i$. We use radial basis function (RBF) interpolation to evaluate sampled grid elements. Given a grid of resolution $N$, the continuous kernel $k_{\text{SO}(n)}$ is evaluated for any $\mathbf{R}$ as:
\begin{align}
    k_{\text{SO}(n)}(\mathbf{R}) = \sum^N_{i=1} a_{d,\psi}(\mathbf{R}, \mathbf{R}_i)\mathbf{k}_i.
\end{align}
Here, $a_{d,\psi}(\mathbf{R}, \mathbf{R}_i)$ represents the RBF interpolation coefficient of $\mathbf{R}$ corresponding to $\mathbf{R}_i$ obtained using Gaussian RBF $\psi$ and Riemannian distance $d$. The uniformity constraint on the grid allows us to scale $\psi$ to the grid resolution dynamically. This ensures that the kernel is smooth and makes our approach hyperparameter-free.

\section{Experiments and Evaluation} \label{sec:results}

In this section, we present our results and evaluation. Section \ref{sec:methodology} introduces our experimental setup. Our results on MedMNIST are presented in Sections \ref{sec:organresults} and  \ref{sec:results1}. Section \ref{sec:results2} offers a deeper look into the generalization performance. Directions for future work based on our results are suggested in Section \ref{sec:futurework}.

\subsection{Evaluation Methodology} \label{sec:methodology}


From here on, we refer to our approach as the SE(3)-CNN. We evaluate the SE(3)-CNNs for different group kernel resolutions. The sample and kernel resolutions are kept equal. We use a regular CNN as our baseline model. We also compare discrete SE(3) subgroup equivariant G-CNNs. K-CNN and T-CNN are equivariant to the 180 and 90-degree rotational symmetries, containing 4 and 12 group elements, respectively. 

All models use the same ResNet \cite{resnet} architecture consisting of an initial convolution layer, two residual blocks with two convolution layers each, and a final linear classification layer. Batch normalization is applied after the first convolution layer. In the residual blocks, we use instance normalization instead. Max spatial pooling with a resolution of $2\times2\times2$ is applied after the first residual block. Global pooling is applied before the final linear layer to produce SE(3) invariant feature descriptors. The first layer maps to 32 channels. The residual blocks map to 32 and 64 channels, respectively. For the G-CNNs, the first convolution layer is a lifting convolution, and the remainders are group convolutions. All spatial kernels have a resolution of $7\times7\times7$. Increasing the group kernel resolution increases the number of parameters. Hence, a second baseline CNN with twice the number of channels is included. The number of parameters of the models is presented in Table \ref{tab:errors}.

We evaluate the degree of SE(3) equivariance obtained by the SE(3)-CNNs on OrganMNIST3D \cite{organmnist1, organminst2} and rotated OrganMNIST3D. For rotated OrganMNIST3D, samples in the test set are randomly rotated. We further evaluate FractureMNIST3D \cite{fracturemnist}, NoduleMNIST3D \cite{nodulemnist}, AdrenalMNIST3D \cite{medmnistv2}, and SynapseMNIST3D \cite{medmnistv2} from the MedMNIST dataset \cite{medmnistv1, medmnistv2}. These volumetric image datasets form an interesting benchmark for SE(3) equivariant methods, as they naturally contain both isotropic and anisotropic features. All input data has a single channel with a resolution of $28\times28\times28$. Each model is trained for 100 epochs with a batch size of 32 and a learning rate of $1\times10^{-4}$ using the Adam \cite{adam} optimizer on an NVIDIA A100 GPU. The results are averaged over three training runs with differing seeds.

\subsection{SE(3) Equivariance Performance} \label{sec:organresults}

\begin{table}[t!]
\centering
\caption{Number of parameters, computational performance, accuracies, and drop in accuracy on test scores between OrganMNIST3D and rotated OrganMNIST3D. Computational performance is measured during training. The highest accuracy and lowest error are shown in \textbf{bold}.}
\label{tab:errors}
\begin{tabular}{lccccccccc}
\toprule
Model & Baseline & Baseline big & \multicolumn{5}{c}{SE(3)-CNN} & K-CNN & T-CNN \\ \midrule
Sample res. & - & - & 4 & 6 & 8 & 12 & 16 & 4 & 12 \\ \midrule
\# parameters & 89k & 200k & 80k & 96k & 111k & 142k & 172k & 80k & 142k \\
Seconds/epoch & 3.37 & 5.83 & 9.67 & 14.43 & 18.48 & 27.31 & 36.37 & 9.70 & 27.75 \\
Memory (GB) & 3.34 & 4,80 & 6.89 & 9.188 & 12.38 & 15.75 & 20.86 & 6.70 & 15.52 \\ \midrule
Accuracy & 0.545 & 0.697 & 0.655 & 0.681 & 0.688 & 0.703 & 0.698 & 0.633 & \textbf{0.722} \\
Rotated Acc. & 0.207 & 0.264 & 0.581 & 0.593 & 0.592 & 0.608 & \textbf{0.628} & 0.327 & 0.511 \\ \midrule
Drop in Acc. \% & 62.15 & 62.09 & 11.26 & 12.91 & 14.07 & 13.60 & \textbf{10.09} & 48.27 & 29.27 \\ \bottomrule
\end{tabular}
\end{table}

Table \ref{tab:errors} shows the accuracies and accuracy drops obtained by the evaluated models on the OrganMNIST3D test set and rotated test set. The decrease in accuracy is calculated as the percentage of the difference between the test scores on the test set and the rotated test set. Both baselines suffer from a high accuracy drop. This is expected, as these models are not equivariant to SE(3). K-CNN and T-CNN fare better. Due to its higher SO(3) kernel resolution, T-CNN outperforms K-CNN. However, these methods do not generalize to the SE(3) group. The SE(3)-CNNs obtain significantly lower drops in accuracy, showing their improved generalization to SE(3). The SE(3)-CNNs at sample resolutions of 12 and 16 also reach higher accuracies than both baseline models. As the sample resolution increases, performance on the standard test set shows a more pronounced increase than on the rotated test set. This results in a slight increase in accuracy drop. At a sample resolution of 16, accuracy on the standard test set decreases while the highest accuracy is obtained on the rotated test set. A high degree of SE(3) equivariance seems disadvantageous on OrganMNIST3D. This would also explain why T-CNN achieved the highest accuracy on the standard test set, as this model generalizes less to SE(3). OrganMNIST3D contains samples aligned to the abdominal window, resulting in high isotropy. This reduces the advantages of SE(3) equivariance.

\subsection{Performance on MedMNIST} \label{sec:results1}

\begin{table}[t!]
\small
\centering
\caption{Accuracies and SE(3)-CNN performance gain in percentage points (p.p.) over the CNN baselines on FractureMNIST3D, NoduleMNIST3D, AdrenalMNIST3D, and SynapseMNIST3D. Sample resolution in parenthesis behind the model name. Standard deviation in parenthesis behind the accuracies. The highest accuracy is indicated in \textbf{bold}.}
\label{tab:medmnist}
\begin{tabular}{lcccc}
\toprule
Model        & Fracture             & Nodule               & Adrenal              & Synapse      \\ \midrule
Baseline     & 0.450 $(\pm 0.033)$          & 0.834 $(\pm 0.019)$           & 0.780 $(\pm 0.006)$          & 0.694 $(\pm 0.006)$ \\
Baseline-big & 0.499 $(\pm 0.032)$          & 0.846 $(\pm 0.010)$          & 0.806 $(\pm 0.022)$          & 0.720 $(\pm 0.019)$ \\ \midrule
SE(3)-CNN (4)    & 0.588 $(\pm 0.029)$          & 0.869 $(\pm 0.008)$          & 0.800 $(\pm 0.006)$          & 0.865 $(\pm 0.010)$ \\
SE(3)-CNN (6)    & 0.617 $(\pm 0.013)$          & \textbf{0.875 $(\pm 0.008)$} & 0.804 $(\pm 0.002)$          & 0.870 $(\pm 0.008)$ \\
SE(3)-CNN (8) & 0.615 $(\pm 0.002)$ & \textbf{0.875 $(\pm 0.014)$} & 0.815 $(\pm 0.015)$ & \textbf{0.885 $(\pm 0.007)$} \\
SE(3)-CNN (12)   & \textbf{0.621 $(\pm 0.002)$} & 0.873 $(\pm 0.005)$          & 0.814 $(\pm 0.010)$          & 0.858 $(\pm 0.028)$ \\
SE(3)-CNN (16)   & 0.604 $(\pm 0.012)$          & 0.858 $(\pm 0.011)$          & \textbf{0.832 $(\pm 0.005)$} & 0.869 $(\pm 0.023)$ \\
\midrule
K-CNN (4) & 0.486 $(\pm 0.012)$ & 0.859 $(\pm 0.013)$ & 0.798 $(\pm 0.011)$ & 0.709 $(\pm 0.009)$ \\
T-CNN (12) & 0.490 $(\pm0.036)$ & 0.862 $(\pm 0.006)$ & 0.800 $(\pm 0.011)$ & 0.777 $(\pm 0.021)$ \\
\midrule
Gain in p.p. & 12.2 & 2.9 & 2.6 & 16.5 \\
\bottomrule
\end{tabular}
\end{table}

The accuracies obtained on FractureMNIST3D, NoduleMNIST3D, AdrenalMNIST3D, and SynapseMNIST3D are reported in Table \ref{tab:medmnist}. The SE(3)-CNNs obtain the highest accuracies on all datasets. On FreactureMNIST3D, the highest accuracy is achieved by the SE(3)-CNN (12). Both K-CNN and T-CNN achieve an accuracy very similar to the baseline models. The baseline-big model slightly outperforms both K-CNN and T-CNN. On NoduleMNIST3D, SE(3)-CNN (6) and SE(3)-CNN (8) achieve the highest accuracy, with SE(3)-CNN (12) performing only slightly lower. K-CNN and T-CNN outperform both baseline models. On AdrenalMNIST3D, the differences in accuracy between all models are the lowest. SE(3)-CNN (16) obtains the highest accuracy, whereas the baseline model obtains the lowest. The baseline-big model outperforms K-CNN and T-CNN. On SynapseMNIST3D, we again observe a significant difference in performance between the SE(3)-CNNs and the other models. SE(3)-CNN (6) obtained the highest performance. T-CNN outperforms K-CNN and both baseline models. However, the baseline-big model outperforms K-CNN. On NoduleMNIST3D and AdrenalMNIST3D, only a slight performance gain is achieved by the SE(3)-CNNs. This is likely due to the isotropy of the samples in these datasets. In these cases, SE(3) equivariance is less beneficial. In contrast, FractureMNIST3D and SynapseMNIST3D are more anisotropic, resulting in significant performance gains of up to 16.5\%.

\subsection{Model Generalization} \label{sec:results2}

\begin{figure}[t!]
     \centering
     \begin{subfigure}[b]{0.49\textwidth}
         \centering
         \includegraphics[width=\textwidth]{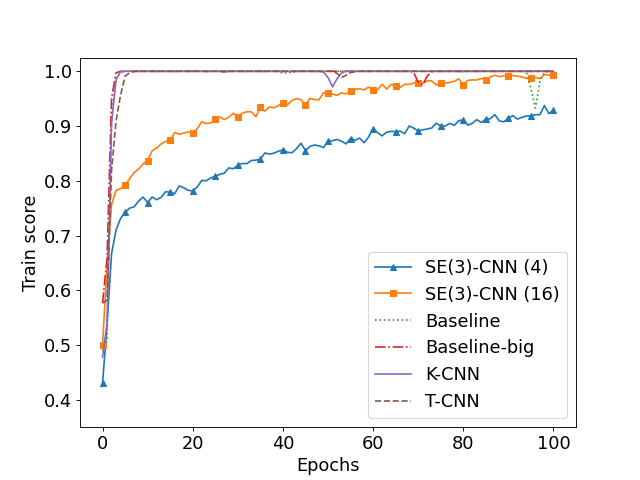}
         \caption{Train scores on SynapseMNIST3D.}
         \label{fig:synapseleft}
     \end{subfigure}\hfill
     \begin{subfigure}[b]{0.49\textwidth}
         \centering
         \includegraphics[width=\textwidth]{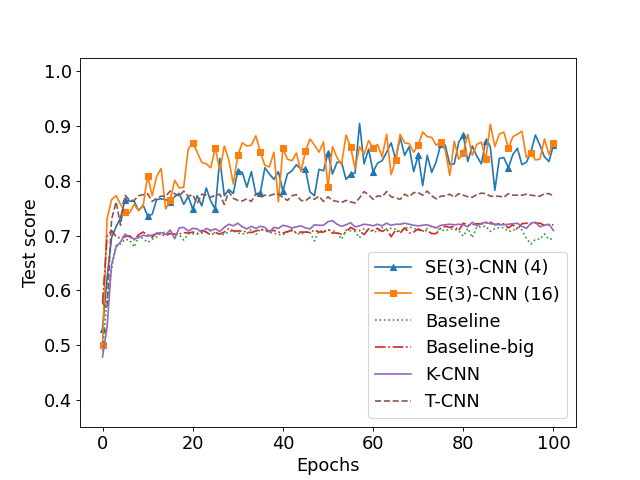}
         \caption{Test scores on SynapseMNIST3D.}
         \label{fig:synapseright}
     \end{subfigure}
     \caption{Accuracy scores of the baseline models, the SE(3)-CNN (4) and (16) models, and the discrete SE(3) subgroup models on (a) the train set and (b) the test set of SynapseMNIST3D.\label{fig:training}}
\end{figure}

The scores obtained on both the train set and test sets of SynapseMNIST3D in Figures \ref{fig:synapseleft} and \ref{fig:synapseright}, respectively. We observed similar behavior on all datasets. Figure \ref{fig:training} shows a stark difference between the SE(3)-CNNs and the other models. The baselines and K-CNN and T-CNN  converge after a few epochs on the train set. SE(3)-CNN (16) requires all 100 epochs to converge on the train set. SE(3)-CNN (4) does not converge within 100 epochs. This improvement during the training window is also observed in the test scores. This suggests that SE(3)-CNNs suffer less from overfitting, which results in improved model generalization. K-CNN and T-CNN behave similarly to the baselines. We hypothesize that this results from the weight-sharing that occurs during the RBF interpolation. We do observe a higher variance in scores of the SE(3)-CNNs, which we attribute to the random nature of the convolution kernels.

\subsection{Future Work} \label{sec:futurework}

With an increase in the sample resolution, a better approximation to SE(3) equivariance is achieved. However, we observe that this does not necessarily improve model performance, e.g., in the case of isotropic features. This could indicate that the equivariance constraint is too strict.
We could extend our approach to learn \textit{partial} equivariance. Rather than sampling on the entire SO(3) manifold, each group convolution layer could learn sampling in specific regions. This suggests a compelling extension of our work, as learning partial invariance has shown to increase model performance \cite{romero2022learning}.
\section{Conclusion} \label{sec:conclusion}

This work proposed an SE(3) equivariant separable G-CNN. Equivariance is achieved by sampling uniform kernels on a continuous function over SO(3) using RBF interpolation. Our approach requires no additional hyper-parameters compared to CNNs. Hence, our SE(3) equivariant layers can replace regular convolution layers. Our approach consistently outperforms CNNs and discrete subgroup equivariant G-CNNs on challenging medical image classification tasks. We showed that 3D CNNs and discrete subgroup equivariant G-CNNs suffer from overfitting. We showed significantly improved generalization capabilities of our approach. In conclusion, we have demonstrated the advantages of equivariant methods in medical image analysis that naturally deal with rotation symmetries. The simplicity of our approach increases the accessibility of these methods, making them available to a broader audience.
\subsection*{Acknowledgments} \label{sec:acknowledgment}

This work was part of the research program VENI with project ”context-aware AI in medical image analysis” with number 17290, financed by the Dutch Research Council (NWO). We want to thank SURF for the use of the National Supercomputer Snellius. For providing financial support, we want to thank ELLIS and Qualcomm, and the University of Amsterdam.

\newpage
\printbibliography

\end{document}